\documentclass[sigconf, screen]{acmart}
\usepackage{xcolor}
\usepackage{adjustbox}
\usepackage{multicol}
\usepackage{amsmath}
\usepackage{multirow}
\usepackage{graphicx}
\usepackage{pbox}
\usepackage{CJKutf8}
\usepackage{tabularx}
\usepackage{threeparttable}
\usepackage{pgf-pie}
\usepackage{graphicx}

\newcommand\red[1]{\textcolor{red}{#1}}

\AtBeginDocument{%
  \providecommand\BibTeX{{%
    \normalfont B\kern-0.5em{\scshape i\kern-0.25em b}\kern-0.8em\TeX}}}

\usepackage{enumitem}
\usepackage{multirow}
\usepackage{array}
\usepackage{bm}

\newcolumntype{P}[1]{>{\centering\arraybackslash}p{#1}}

\copyrightyear{2025}
\acmYear{2025}
\setcopyright{acmlicensed}\acmConference[MM '25]{Proceedings of the 33rd ACM International Conference on Multimedia}{October 27--31, 2025}{Dublin, Ireland}
\acmBooktitle{Proceedings of the 33rd ACM International Conference on Multimedia (MM '25), October 27--31, 2025, Dublin, Ireland}
\acmDOI{10.1145/3746027.3758289}
\acmISBN{979-8-4007-2035-2/2025/10}

\begin{document}

\title{HateClipSeg: A Segment-Level Annotated Dataset for Fine-Grained Hate Video Detection}


\author{Han Wang}
\orcid{0009-0007-4486-0693}
\affiliation{
  \institution{Singapore University of Technology and Design}
  \streetaddress{8 Somapah Road}
  \country{Singapore}
  \postcode{487372}
}
\email{han_wang@mymail.sutd.edu.sg}
\authornote{Both authors contributed equally. Authors are listed in alphabetical order.}

\author{Zhuoran Wang}
\orcid{0009-0002-9635-5182}
\affiliation{
  \institution{Singapore University of Technology and Design}
  \streetaddress{8 Somapah Road}
  \country{Singapore}
  \postcode{487372}
}
\email{zhuoran_wang@sutd.edu.sg}
\authornotemark[1]

\author{Roy Ka-Wei Lee}
\orcid{0000-0002-1986-7750}
\affiliation{
  \institution{Singapore University of Technology and Design}
  \streetaddress{8 Somapah Road}
  \country{Singapore}
  \postcode{487372}
}
\email{roy_lee@sutd.edu.sg}





\begin{abstract}
Detecting hate speech in videos remains challenging due to the complexity of multimodal content and the lack of fine-grained annotations in existing datasets. 
We present \textsf{HateClipSeg}, a large-scale multimodal dataset with both video-level and segment-level annotations, comprising over 11,714 segments labeled as \textit{Normal} or across five \textit{Offensive} categories: \textit{Hateful}, \textit{Insulting}, \textit{Sexual}, \textit{Violence}, \textit{Self-Harm}, along with explicit target victim labels. Our three-stage annotation process yields high inter-annotator agreement (Krippendorff's alpha = 0.817). We propose three tasks to benchmark performance: (1) \textit{Trimmed Hateful Video Classification}, (2) \textit{Temporal Hateful Video Localization}, and (3) \textit{Online Hateful Video Classification}. Results highlight substantial gaps in current models, emphasizing the need for more sophisticated multimodal and temporally aware approaches. The HateClipSeg dataset are publicly available at \url{https://github.com/Social-AI-Studio/HateClipSeg.git}.

\red{\textbf{Disclaimer: This paper contains sensitive content that may be disturbing to some readers.}}
\end{abstract}



\maketitle 



\section{Introduction}

The growing prevalence of online hate speech poses significant societal challenges, especially as multimodal content, combining text, visuals, and audio, enhances its reach and subtlety~\cite{hee2024recent}. Unlike unimodal forms, multimodal hate speech leverages cross-modal interactions to mask or amplify harmful messages, making detection more difficult (e.g., benign text paired with violent imagery or sarcastic tone). Recent efforts, such as HateMM~\cite{das2023hatemm} and MultiHateClip~\cite{wang2024multihateclip}, highlight increasing research attention. HateMM includes 1,083 BitChute videos labeled as hateful or normal, while MultiHateClip features 2,000 YouTube and Bilibili videos annotated as hateful, offensive, or normal.


Nevertheless, existing hate video datasets and detection methods are limited in supporting nuanced content moderation~\cite{wang2025crossmodal}. Most use coarse video-level labels (e.g., hateful vs. normal), which obscure specific hate types. Although some provide hate speech locations, segment annotations rely on annotators’ subjective boundary decisions, making quality difficult to measure and ensure. However, fine-grained prediction enables more transparent and differentiated moderation by identifying specific hate types and targets (e.g., racial slurs vs. sexual insults), facilitating nuanced policy enforcement and fairer appeals. Segment-level understanding also offers richer supervision for models with temporal and contextual awareness. In practice, moderators often need to remove only hateful segments rather than entire videos, balancing policy enforcement with user rights. In live-streaming contexts, real-time detection allows prompt flagging and intervention. Without high-quality, fine-grained segment annotations, systems risk over-moderation (i.e., removing entire videos for limited harm) or under-moderation (i.e., missing subtle hate), undermining both safety and free expression.

To address these limitations, we introduce \textsf{HateClipSeg}, a multimodal dataset with fine-grained, segment-level annotations for hate video detection. Our goal is to bridge the gap between coarse video-level labels and the real-world need for precise, temporally localized detection of nuanced, context-dependent hate speech. \textsf{HateClipSeg} advances multimodal hate speech research by enabling reliable detection of \textit{Offensive} segments, further categorized into \textit{Hateful}, \textit{Insulting}, \textit{Sexual}, \textit{Violence}, and \textit{Self-Harm} across multiple modalities.

Our research objectives are threefold. First, we aim to establish a benchmark dataset that provides predefined, semantically coherent segment boundaries, addressing the challenge of inconsistent annotator-determined segmentation and ensuring consistent, reproducible annotations. Second, we design a three-step annotation pipeline, \textit{independent annotation}, \textit{paired discussion}, and \textit{re-annotation}, that significantly improves inter-annotator agreement, achieving a video-level \textit{Offensive} or \textit{Normal} Krippendorff's alpha of 0.817. 
This approach addresses the inherent ambiguity of multimodal content and ensures high-quality labels across all types of annotation. Third, we demonstrate the practical utility of \textsf{HateClipSeg} by benchmarking state-of-the-art models across three challenging tasks: \textit{trimmed hateful video classification}, \textit{temporal hateful video localization}, and \textit{online hateful video detection}. Our experiments reveal that while current models perform moderately in trimmed video classification (69.48 Macro-F1), their performance drops sharply in temporal localization (29.42 F1 at tIoU=0.7) and remains limited in online classification (62.75 Macro-F1), underscoring the pressing need for more advanced, multimodal, and temporally aware hate speech detection systems. 

We summarize our contributions as follows: (i) We present \textsf{HateClipSeg}, a multimodal hate speech dataset containing both video- and segment-level annotations across five fine-grained \textit{Offensive} categories, and explicit target-victim labels. (ii) We introduce a three-step annotation pipeline that significantly improves inter-annotator agreement, ensuring high-quality labels across all types of annotation. (iii) We demonstrate the dataset’s broad applicability through three challenging tasks: Trimmed Video Classification, Temporal Localization, and Online Classification, and provide comprehensive benchmarks that reveal substantial performance gaps in current state-of-the-art models.

\section{Dataset Construction}

This section presents our pipeline for constructing \textsf{HateClipSeg}. We describe the processes of data collection and processing, annotation protocols, quality analysis, and dataset statistics, highlighting how \textsf{HateClipSeg} overcomes key limitations of prior datasets.

\begin{figure}[t]
  \centering
  \includegraphics[width=0.90\linewidth]{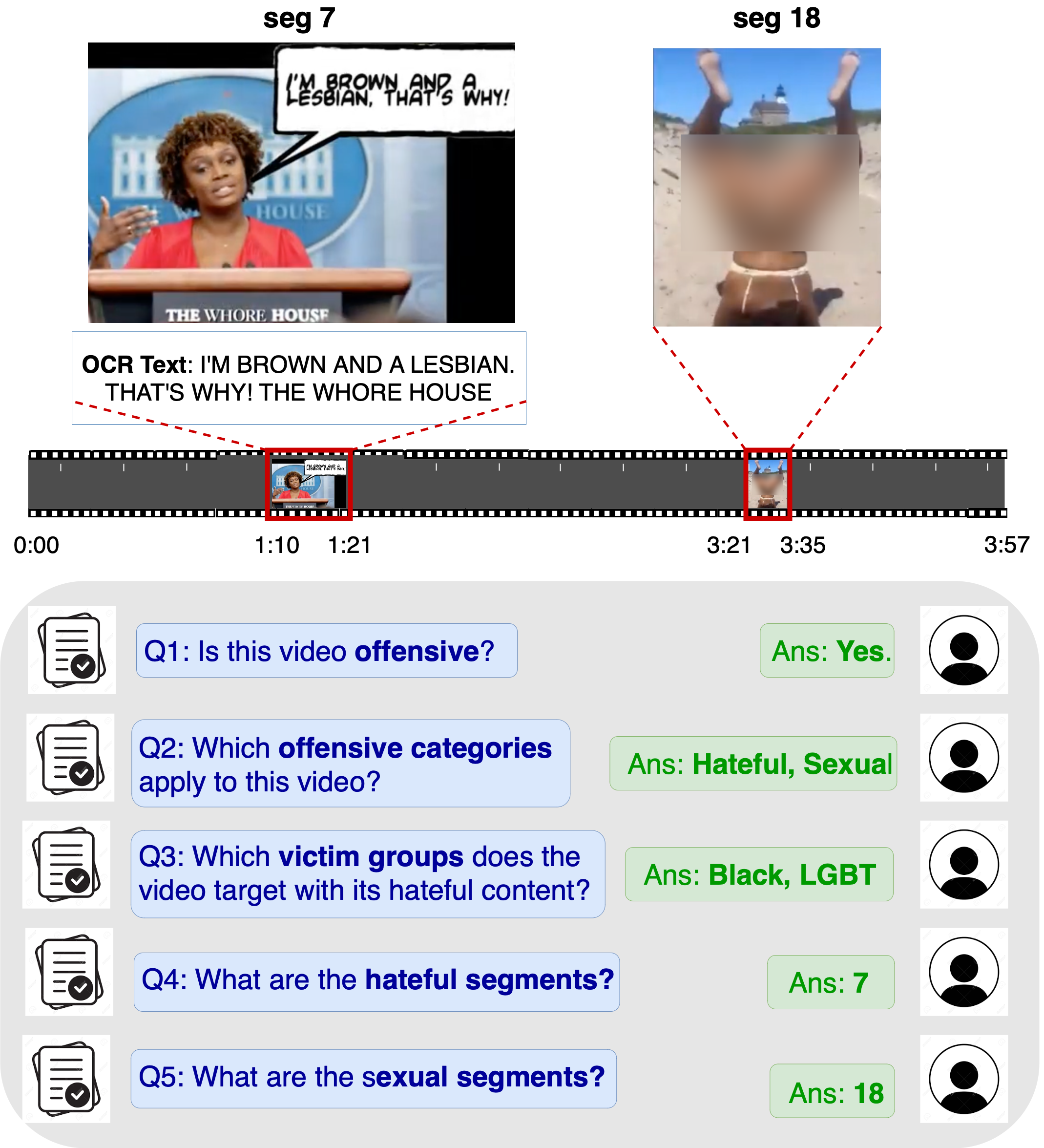}
  \caption{An example of a video clip annotated by annotators.}
  \label{fig:annotation-example}
\end{figure}

\subsection{Data Collection and Processing}

We compiled a lexicon of over 100 frequently used terms and phrases targeting four common categories of hate speech: \textit{race}, \textit{gender}, \textit{religion}, and \textit{sexuality}. The lexicon was developed using entries from established resources such as Hatebase~\cite{Davidson2017automated} and HateXplain~\cite{mathew2021hatexplain}, supplemented with additional terms manually curated by our team.\footnote{The final lexicon will be publicly released with the dataset to support reproducibility and future research.} Below are examples of selected entries from the lexicon:

\begin{itemize}
\item \textbf{Race}: wetback, ching chong, nig**r
\item \textbf{Gender}: cu*t, whore, incel
\item \textbf{Religion}: goatf***er, kike, dothead
\item \textbf{Sexuality}: fa*got, dyke, tranny
\end{itemize}

Using the constructed hate lexicons, we performed keyword searches on YouTube and BitChute, a minimally moderated site known for hosting extremist and conspiratorial content~\cite{trujillo2021bitchute}. Videos were limited to durations between 3 and 10 minutes, yielding an initial dataset of 4,745 videos.

To reduce annotation costs and increase the proportion of hateful content, we employed a pretrained model to filter out non-hateful videos. Following the video fine-tuning strategy in~\cite{wang2025crossmodal}, we fine-tuned LLaMA-3.2-11B~\cite{dubey2024llama} on the MultiHateClip dataset~\cite{wang2024multihateclip}, which labels videos as hateful, offensive, or normal. We retained only those predicted as hateful, resulting in a candidate pool of 435 videos.
To enable segment-level annotation, we automatically divided videos into semantically coherent segments, each representing a complete semantic unit. First, Whisper~\cite{radford2022robustspeechrecognitionlargescale} generated transcripts with word-level timestamps, which were merged into sentence-level segments using the NLTK~\cite{bird-loper-2004-nltk} Punkt tokenizer. Silent intervals longer than 20 seconds were further segmented at scene changes, identified by drops in cosine similarity between consecutive ViT-based frame embeddings. Manual inspection of 250 segments from 10 diverse videos showed that 90\% captured self-contained content (e.g., complete sentences or coherent scenes). This process yielded 11,714 segments with an average length of 8.84 seconds.

\subsection{Annotation Question}

For each video, annotators first assigned a primary video-level label (either \textit{Offensive} or \textit{Normal}). \textit{Normal} content was defined as material not falling into any of these offensive categories. If a video was labeled \textit{Offensive}, annotators further specified the type of offense by selecting one or more categories from \textit{Hateful}, \textit{Insulting}, \textit{Sexual}, \textit{Violence}, or \textit{Self-Harm}. Specifically, \textit{Hateful} refers to content expressing or inciting hatred or violence against protected groups, \textit{Insulting} includes demeaning or dehumanizing language, \textit{Sexual} involves explicit sexual content or pornography, \textit{Violence} covers depictions or glorification of physical harm, and \textit{Self-Harm} denotes content promoting self-injury or suicide.

For each video labeled as \textit{Offensive}, annotators identified the specific segments where the offensive content appeared. Additionally, for videos labeled \textit{Hateful}, they selected one or more target victim groups from a predefined list. The predefined victim groups, which include \textit{Jewish}, \textit{LGBT}, \textit{Black}, \textit{White}, \textit{Woman}, \textit{Islam}, among others, were selected based on common targets of hate speech identified in prior literature~\cite{schmidt2017survey} and public discourse. To ensure flexibility, annotators could input additional victim groups under the ``\textit{Other}'' category if the target was not listed. These unlisted groups were later aggregated to identify potential new categories for future evaluation and research. To illustrate our multi-label segment annotation approach, Figure~\ref{fig:annotation-example} shows snapshots from a video labeled with both \textit{Hateful} and \textit{Sexual} categories, targeting \textit{Black} and \textit{Lesbian}.


\begin{table}[t]
\centering
\small
\caption{Krippendorff's alpha inter-annotator agreement scores for each annotation task before and after discussion. O:offensive, N:normal}
\begin{tabular}{lcc}
\toprule
\textbf{Annotation Task} & \textbf{Before Discussion} & \textbf{After Discussion} \\
\midrule
Video-level O/N Label     & 0.791 & 0.817 \\
Segment-level O/N Label      & 0.715 & 0.757 \\
Offensive Category Label      & 0.840 & 0.899 \\
Target Victim Label     & 0.716 & 0.721 \\
\bottomrule
\end{tabular}
\label{tab:iaa_scores}
\end{table}

\begin{table}[t]
  \centering
  \small
  \caption{Video-level and Segment-level Label Distribution.}
  \begin{tabular}{lrr}
    \toprule
    Label       & Video Count  & Segment Count \\
    \midrule
    Hateful      & 194  & 2363       \\
    Insulting    & 280  & 2920      \\
    Sexual        & 69 &   372    \\
    Violent      & 192  & 1281    \\
    Self-Harm    & 18  &    39   \\
    \midrule
    Offensive\textsuperscript{*} & 380 & 5223 \\
     Normal                      & 55 & 6491    \\
    \bottomrule
  \end{tabular}
  \label{tab:annotation_stats}
\end{table}


\subsection{Data Annotation}

We recruited ten undergraduate native English speakers (ages 18–24) as annotators, each completing a training session with 30 example videos. However, training alone did not minimize disagreements. Because each video lasts 3 to 10 minutes and contains numerous segments. This occasionally led to segments being overlooked or misinterpreted, contributing to annotation inconsistencies. To address this, we implemented pairwise discussions, enabling annotators to collaboratively review and reconcile their annotations, clarify interpretations, and ensure that overlooked segments were properly identified. As a result, we organized annotators into pairs and adopted a three-stage annotation process:

\begin{enumerate}
\item Each annotator independently answered all questions.
\item Pairs discussed their annotations to resolve disagreements.
\item Annotators submitted a second round of individual annotations for unresolved disagreements.
\end{enumerate}

To reduce potential bias during discussions, we required annotators to first provide independent annotations. During pairwise discussions, annotators followed a structured protocol focused on objective evidence from the video (e.g., timestamps, visual/audio cues) rather than subjective impressions. They were also encouraged to avoid dominant opinions and evaluative language. For each video, pairs systematically reviewed disagreements, aiming for consensus based on explicit evidence. In cases lacking consensus (approximately 5.3\% videos in \textit{Offensive} or \textit{Normal} annotation), a neutral third annotator independently reviewed the video without access to prior labels. The final \textit{Offensive} or \textit{Normal} label was assigned based on majority voting. For videos labeled as \textit{Offensive}, offensive category, target victim and segment annotations were consolidated accordingly.


We evaluated the effectiveness of our annotation protocol using Krippendorff’s alpha, which is suitable for multi-label data. As shown in Table~\ref{tab:iaa_scores}, the results indicate substantial improvements in inter-annotator agreement across all tasks, with alpha values exceeding 0.8 for both the Video-level offensive (O)/ normal (N) Label and Video-level Offensive Category Label after discussion.
Notably, our inclusion of segment-level inter-annotator agreement provides a level of granularity absent in prior hateful-video datasets, which only report video-level agreement. By validating annotations at both the video and segment level, we enhance transparency and reliability, supporting more robust online hate detection and temporal localization.

Given the potentially harmful nature of the content, annotators were explicitly warned about its offensive nature before participating. They were also provided access to psychological support resources through the university. To prioritize well-being, annotators were encouraged to take breaks during annotation sessions and were allowed to opt out of reviewing any video without penalty.





\begin{figure}[t]
  \centering
  \includegraphics[width=0.7\linewidth]{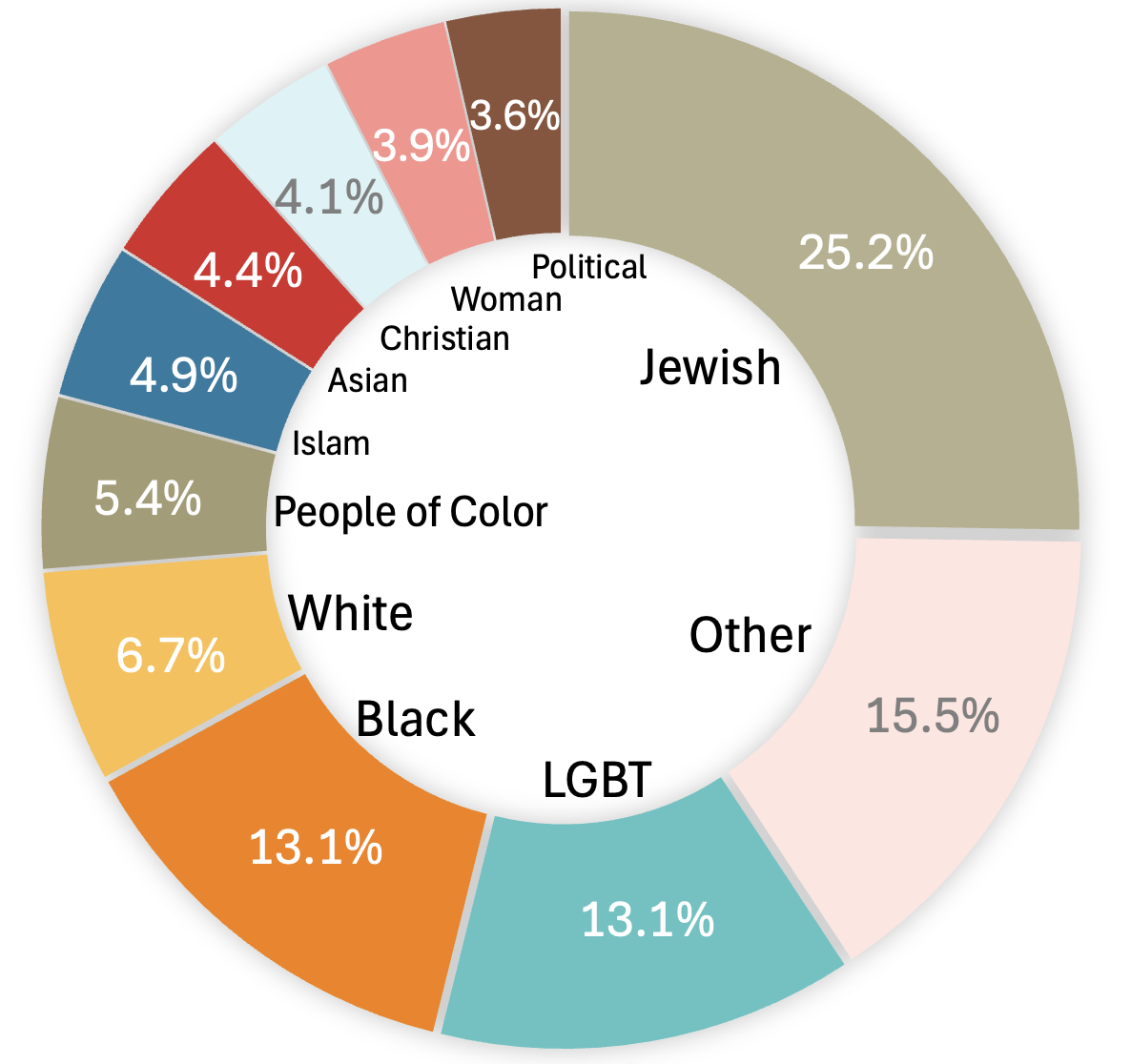}
  \caption{Target victim distribution (video‐level).} 
  \label{fig:victim_pie_chart}
\end{figure}

\subsection{Data Analysis}


\textsf{HateClipSeg} consists of 435 videos and a total of 11,714 segments, making it a comprehensive resource for multimodal hate speech analysis. The segments have an average of 8.84 seconds, providing a fine-grained temporal structure for detailed content analysis. Each segment contains an average of 17.7 words, reflecting diverse levels of linguistic complexity.

As shown in Table~\ref{tab:annotation_stats}, 87\% of videos contain at least one \textit{offensive} segment, demonstrating the effectiveness of our non-hate video filtering strategy. At the segment level, the distribution is more balanced, with 5223 \textit{offensive} and 6491 \textit{normal} segments, indicating that even videos labeled as \textit{offensive} include many \textit{normal} segments. This balance supports both binary and fine-grained temporal classification tasks. Offensive content is distributed across five categories—\textit{Hateful}, \textit{Insulting}, \textit{Sexual}, \textit{Violence}, and \textit{Self-Harm}—with \textit{Insulting} being the most frequent (2920 segments), followed by \textit{Hateful} (2363 segments). The \textit{Offensive} label consolidates these categories, and due to overlapping labels, the sum of individual category counts exceeds the total number of offensive segments.


The victim group distribution, illustrated in Figure~\ref{fig:victim_pie_chart}, highlights the diversity of targeted groups in \textsf{HateClipSeg}. "The 'Other' category (15\%) captures victim identities not covered by predefined labels. Among specific groups, \textit{Jew} (25\%) is the most frequently targeted, followed by \textit{Black}, \textit{LGBT}, and \textit{White}. The remaining categories—\textit{People of Color}, \textit{Islam}, \textit{Asian}, \textit{Woman}, and \textit{Christian}—reflect diverse racial, religious, and gender-based hate. This diversity underscores the dataset’s capacity to support research on both common and less explicit forms of targeted hate.

Compared to existing datasets such as HateMM and MultiHateClip, \textsf{HateClipSeg} offers a larger scale, finer granularity at the segment level, and specific offensive category. This richness supports the development and evaluation of models for nuanced, multimodal hate speech detection and localization.

\section{Task Formulation}

Previous hate video detection research has primarily focused on video-level classification~\cite{das2023hatemm,wang2024multihateclip,wang2025crossmodal}. In contrast, \textsf{HateClipSeg} provides high-quality segment-level annotations, enabling more fine-grained analysis of multimodal content. This supports the formulation of three complementary tasks that reflect real-world challenges in content moderation:


\begin{enumerate}
    \item \textbf{Trimmed Video Classification}: Predict a single label for each pre-segmented clip.
    \item \textbf{Temporal Video Localization}: Detect labels along with their start and end timestamps within untrimmed videos.
    \item \textbf{Online Video Classification}: Perform real-time label prediction on streaming video.
\end{enumerate}

These tasks leverage the unique properties of \textsf{HateClipSeg}, advancing hate detection beyond traditional video-level classification. Detailed definitions and formal problem statements follow.


\begin{figure}[t]
  \centering
  \includegraphics[width=0.9\linewidth]{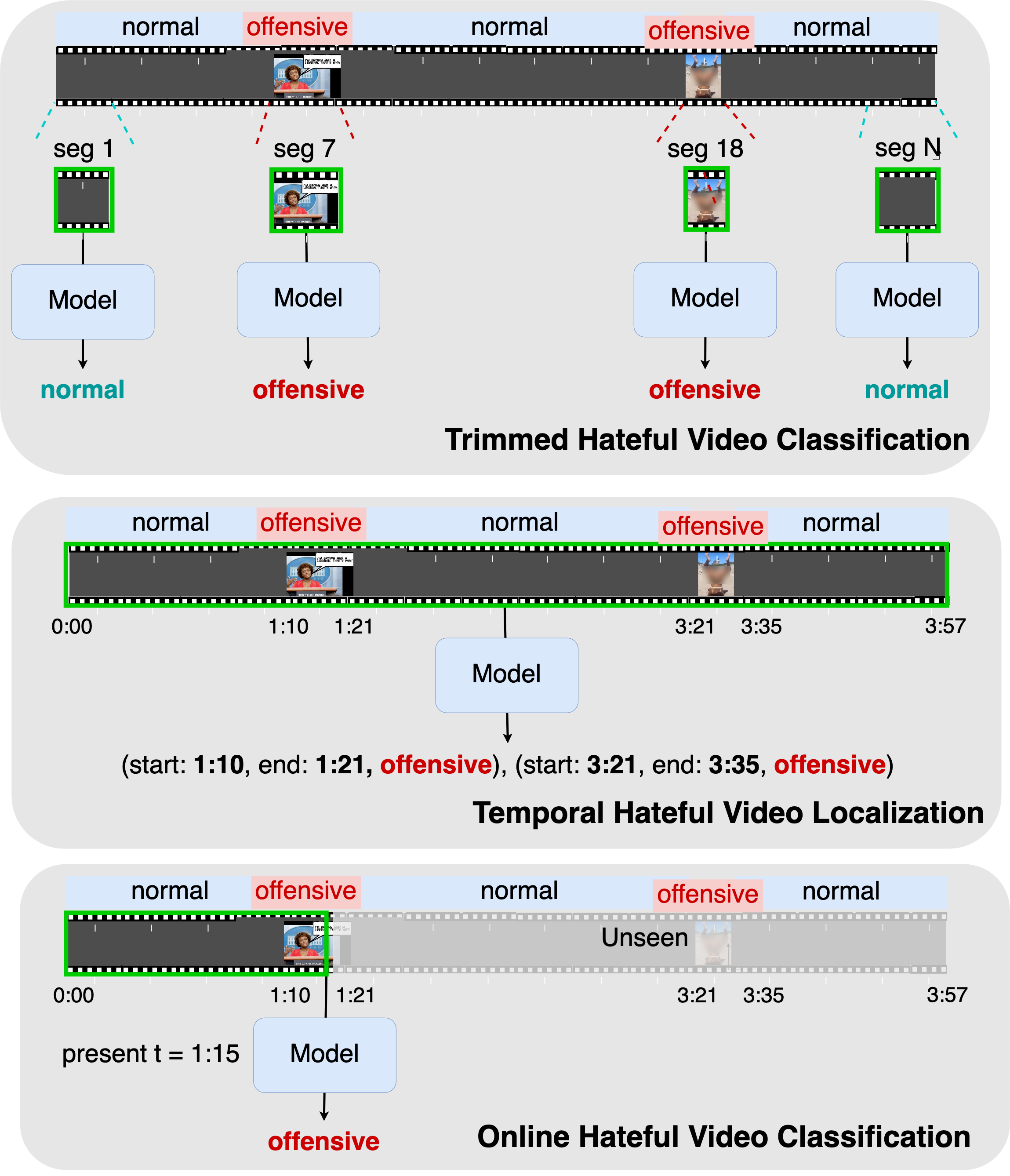}
  \caption{Diagram illustrating Trimmed Hateful Video Classification, Temporal Hateful Video Localization, and Online Hateful Video Classification}
  \label{fig:tasks_diagram}
\end{figure}


\subsection{Trimmed Hateful Video Classification}

Trimmed video classification assigns a label to a pre-segmented clip that is temporally self-contained and semantically complete. This task serves as a baseline, simplifying the challenge by removing the need for temporal boundary detection. Conceptually, it is comparable to the hateful video classification problem studied in prior work, where each input clip is treated as a whole, and the model predicts whether it contains offensive content. This formulation aligns with practical scenarios where video segments are pre-extracted or provided in isolation.

Let a trimmed video segment be represented by a multimodal feature tuple \( x = (x^v, x^t, x^a) \), where \( x^v \), \( x^t \), and \( x^a \) denote visual, textual, and audio features, respectively. For consistency, all segments originally labeled as \textit{hateful}, \textit{insulting}, \textit{sexual}, \textit{violent}, or \textit{self-harm} are merged into a single \textit{offensive} category, resulting in a binary classification scheme used throughout all downstream tasks.


The objective is to learn a classification function \( f \) such that:
\[
f(x) = y, \quad y \in \mathcal{Y},
\]
where \( y = 1 \) denotes \textit{offensive} and \( y = 0 \) denotes \textit{normal}.

Leverage the strong capabilities of large language models in hate speech detection \cite{wang2023evaluating,nirmal2024interpretable,chiu2021gpt3}. We follow the experimental setup of~\cite{wang2025crossmodal}, which achieved state-of-the-art performance in hate video detection using the LLaMA-3.2-11B vision-language model (VLM). This model supports both unimodal and multimodal (vision-text) inputs and is fine-tuned on \textsf{HateClipSeg}. Due to current limitations in large models, audio features are excluded.





\subsection{Temporal Hateful Video Localization}
Temporal localization seeks to identify labeled segments with precise start and end timestamps within untrimmed videos. This task reflects the real-world challenge of detecting offensive content embedded in long-form videos where the boundaries of harmful content are not pre-defined. Given an untrimmed video represented by a feature sequence \( \mathbf{x} = [x_1, x_2, \ldots, x_T] \), where \( x_{t_i} \) is the feature at time \( t_i \), the task is to predict a set of \textit{offensive} segments:
\[
g(\mathbf{x}) = \{(s_j, e_j, y_j = 1)\}_{j=1}^M, \quad 0 \leq s_j < e_j \leq T,
\]
where \( M \) is the number of proposed candidates, and \( s_j \) and \( e_j \) denote the start and end timestamps of the \( j \)-th \textit{offensive} segment.

Prior research on video understanding has developed Temporal Action Localization (TAL) techniques to detect actions and their temporal boundaries in untrimmed videos \cite{yang2020revisiting}. While TAL has not been applied to hateful video detection, we hypothesize that its ability to localize sparse, temporally constrained events is transferable to our task of segmenting offensive content. TAL models are designed to handle long video sequences with imbalanced foreground (action) and background (non-action) segments—conditions that closely resemble our dataset, where offensive segments are sparse and embedded within mostly normal content.

Based on this reasoning, we adopt ActionFormer~\cite{zhang2022actionformer}, a state-of-the-art Transformer-based model for temporal action localization. ActionFormer integrates multiscale feature representations with local self-attention and a lightweight decoder to classify each temporal moment $t_i$ and predict action boundaries. Since our dataset includes only a single foreground class (\textit{offensive}), with \textit{normal} treated as background, this task tests a model’s ability to localize sparse offensive segments within lengthy, mostly normal videos. ActionFormer supports only unimodal visual features; thus, we conduct unimodal experiments and derive multimodal results via late fusion of unimodal predictions.

\subsection{Online Hateful Video Classification}
Online classification involves developing a model capable of monitoring a streaming video and predicting labels in real time. Unlike the temporal localization task, the model cannot observe future frames and must base predictions solely on past and current input. This constraint mimics the challenge of detecting offensive content in live-streaming platforms and social media.

Assume a temporally aligned feature sequence \( x = [x_1, x_2, \ldots, x_T] \), where \( x_{t_i} \) represents the feature at time \( t_i \). The model typically maintains a maximum context window of size \( N \), accessing features in the interval \( [t_{i-N}, t_i] \), denoted as \( x_{t_{i-N}}^{t_i} \). The goal is to predict whether the current moment is offensive or normal:
\[
h\big(x_{t_{i-N}}^{t_i}\big) = y_{t_i}, \quad y \in \mathcal{Y}, \quad t_i = N, N+s, \ldots, T,
\]
 \( s \) is the stride, and \( y_{t_i} \) indicates \textit{offensive} (1) or \textit{normal} (0) at time \( t_i \).

Online action recognition has been extensively studied in the field of human action understanding, focusing on identifying ongoing actions in streaming data \cite{wang2021oadtr,xu2021long,yang2022colar}. While these techniques have not been directly applied to hateful video detection, we posit that their ability to process streaming sequences and make real-time predictions is well-suited to our task. In both scenarios, the model must infer the presence of a target category (e.g., action or offensive content) based on partial, sequential input, often in the face of ambiguity and limited context.

Based on these parallels, we adopt the Long Short-Term Transformer (LSTR)~\cite{xu2021long} as our baseline model. LSTR uses an encoder to capture long-term dependencies and a decoder for short-term context, effectively extending the model’s temporal receptive field. Since LSTR supports only unimodal visual inputs, we perform unimodal experiments directly and derive multimodal results via late fusion of unimodal predictions.

\section{Experiment}

This section details the use of \textsf{HateClipSeg} across the three tasks introduced in the Task Formulation section, demonstrating its broad applicability and highlighting the inherent challenges of each task.


\subsection{Data Pre-processing}
For the trimmed video classification task, we employ a VLM that accepts raw images and text. Specifically, we sample a single representative frame from each trimmed segment and extract the corresponding transcript using the Whisper model~\cite{radford2022robustspeechrecognitionlargescale}. 

For temporal localization and online classification, temporally aligned, modality-specific features are required. We encode visual features using a frozen ViT-Large~\cite{dosovitskiy2020vit} at each timestamp $t_i$. Text features are extracted using a frozen BERT-Base~\cite{devlin2019bert}, encoding words from $[t_{i-n}, t_i]$ with $n=2$ seconds. Audio features are extracted using frozen Wav2Vec-Emotion~\cite{wagner2023dawn} over $[t_{i-n}, t_i]$ with $n=4$ seconds. Encoder selection and window sizes were empirically optimized based on validation performance.


\subsection{Experiments Setting}
\textbf{Training.} The dataset is split into training (80\%) and testing (20\%). For trimmed video classification, we fine-tune LLaMA-3.2-11B~\cite{dubey2024llama} using LoRA, applying three configurations: visual-only, text-only, and vision-text fusion. Models are trained for 10 epochs, and the best Macro-F1 score is reported. For temporal localization, we implement We adopt ActionFormer~\cite{zhang2022actionformer}, training separate models for each modality. As visual features yield the strongest unimodal performance, we apply late fusion by aligning non-visual predictions to their nearest visual counterparts. Segment boundaries are averaged, and labels assigned via majority voting. Following ActionFormer's design of one candidate per moment, we define the moment rate as 4 FPS and train for 30 epochs.
For online classification, we use LSTR~\cite{xu2021long} as the baseline, training separate models for each modality and fusing predictions by majority voting at each timestamp. The context window is set to 32s with a stride of 0.25s. Models are trained for 5 epochs.

\textbf{Evaluation Metrics.} We report Accuracy, Macro-F1, and class-wise Precision, Recall, and F1 for the offensive class. For temporal video localization, we use temporal Intersection over Union (tIoU) between predicted and ground truth segments, reporting performance at tIoU thresholds of 0.3, 0.5, and 0.7. Since only offensive segments are predicted, macro metrics are not applicable;
we report Accuracy, Precision, Recall, and F1 for the offensive class. For online video classification, we treat each timestamp prediction as an independent data point, applying standard metrics.

\subsection{Results of Trimmed Hateful Video Classification}

Table~\ref{tab:trimmed_classification} shows that the multimodal (V+T) model achieves the highest Macro-F1 score of 69.48, outperforming both text-only and visual-only configurations. This highlights the importance of combining modalities, as textual cues alone outperform visual-only models, indicating their stronger role in detecting offensive content.

However, the performance lags behind prior benchmarks, where LLaMA-3.2-11B achieved Macro-F1 scores of 0.78 on MultiHateClip~\cite{wang2024multihateclip} and 0.81 on HateMM~\cite{das2023hatemm} in~\cite{wang2025crossmodal}. This gap likely stems from two factors: (1) our dataset's greater scale and complexity, with 11,714 segment-level samples versus around 1,000 video-level samples in prior datasets; and (2) the intrinsic difficulty of segment-level classification, as short clips often lack broader context needed to disambiguate subtle or implicit hate speech. These results emphasize the challenges of segment-level classification and the need for models that can better capture multimodal cues and contextual dependencies, even in isolated, short segments.


\begin{table}[t]
\centering
\small
  \caption{Model performance for trimmed hateful video classification. LLaMA: LLaMA-3.2-11B, O: offensive, Acc: accuracy, M-F1: macro-F1, R: recall, P: precision.}
  \begin{tabular}{cccccccc}
    \toprule
     \textbf{Model}&\textbf{Modality}    & \textbf{Acc}  &\textbf{M-F1} & \textbf{F1(O)} & \textbf{R(O)}  & \textbf{P(O)}  \\
    \midrule
\multirow{3}{*}{LLaMA}  & \textit{V} & 59.63  &57.56  & 48.19 & 40.09 & 60.39 \\ 
 & \textit{T} &  64.83 & 62.92 & 54.51 & 45.00 & 69.13\\ 
 & \textit{V, T} & 69.64 & 69.48 & 67.26 & 66.57 & 67.96\\  
    \bottomrule
  \end{tabular}
 \label{tab:trimmed_classification}
\end{table}

\subsection{Results of Temporal Hateful Video Localization}

\begin{table}[t]
\centering
\small
\caption{Model performance for temporal hateful video localization at different tIoU thresholds. O: offensive, Acc: Accuracy, R: Recall, P: Precision.}
\begin{tabular}{cccccccc}
  \toprule
  \textbf{tIoU} & \textbf{Modality}  & \textbf{Acc} & \textbf{F1(O)} &  \textbf{R(O)} & \textbf{P(O)} \\
  \midrule
  0.30 & V  & 42.22 & 59.38 &  84.74 & 45.70 \\
  0.30 & T  & 28.61 & 44.49 &  81.06 & 30.66 \\
  0.30 & A  & 25.16 & 40.21 &  83.74 & 26.45 \\
  0.30 & V, T, A  & 41.83 &  58.98 & 84.18 & 45.40 \\
  0.50 & V  & 35.73 & 52.65 &  75.14 & 40.52 \\
  0.50 & T  & 20.92 & 34.60 &  63.04 & 23.84 \\
  0.50 & A  & 20.80 & 25.40 & 71.73 & 22.66 \\
  0.50 & V, T, A  & 34.16 &  50.92 & 72.68 & 39.19 \\
  0.70 & V   & 18.34 & 30.99 &  44.23 & 23.85 \\
  0.70 & T   & 9.44 & 11.89 &  31.43 & 17.25 \\
  0.70 & A  & 10.39 & 18.83 &  39.21 & 12.39  \\
  0.70 & V, T, A   & 17.24&  29.42 & 41.98 & 22.64 \\
  \bottomrule
\end{tabular}
\label{tab:tiou_metrics}
\end{table}

Table~\ref{tab:tiou_metrics} indicates that the visual modality achieves the highest F1 score of 59.38 at tIoU=0.3, but its performance drops sharply to 30.99 at tIoU=0.7. Late fusion fails to improve multimodal results, underscoring the challenge of accurately localizing short and subtle segments of offensive content.
Temporal hateful video localization presents greater challenges than traditional action localization tasks due to the subtle, context-dependent nature of hate speech, which often lacks clear visual or audio cues. Offensive content is frequently intertwined with benign segments, making boundary delineation ambiguous. The fine-grained, segment-level annotations in \textsf{HateClipSeg} further demand high model sensitivity. These findings reveal the limitations of current temporal models such as ActionFormer in handling fine-grained, multimodal hate localization, highlighting the need for architectures that more effectively capture cross-modal and temporal context.


\subsection{Results of Online Hateful Video Classification}

Table~\ref{tab:online_classification} shows that the multimodal approach achieves the highest Macro-F1 score of 62.75, outperforming unimodal models. Audio features alone perform better than visual and text inputs, reflecting the role of prosody in conveying hate speech. The text-only model underperforms, likely due to many silent or non-speech segments in the dataset. The performance drop from 69.48 (trimmed) to 62.75 (online) underscores the challenge of real-time prediction. Unlike trimmed clips with full context, streaming data requires the model to infer intent from partial information, where offensive cues may be subtle and distributed across modalities. These results emphasize the need for models that can dynamically integrate multimodal cues and handle streaming constraints, potentially via memory-augmented or continual learning approaches.


\begin{table}[t]
\centering
\small
  \caption{Model performance for online hateful video classification. O: offensive, Acc: accuracy, M-F1: macro-F1, R: recall, P: precision.}
  \begin{tabular}{ccccccc}
    \toprule
     \textbf{Model}&\textbf{Modality}    & \textbf{Acc}  &\textbf{M-F1} & \textbf{F1(O)} & \textbf{R(O)}  & \textbf{P(O)}  \\
    \midrule
\multirow{4}{*}{LSTR} & \textit{V}  & 57.99 & 57.52 & 62.00 & 70.33 & 55.43 \\  
 & \textit{T}  & 58.86 & 56.51 & 46.40 & 36.55 & 63.51\\  
 & \textit{A}  & 61.05 & 60.84 & 57.96 & 55.12 & 61.11\\  
 & \textit{V, T, A} & 63.21 & 62.75 & 58.59 & 53.42 & 64.86\\  
    \bottomrule
  \end{tabular}
 \label{tab:online_classification}
\end{table}

\section{Conclusion}

In this paper, we introduced \textsf{HateClipSeg}, a large-scale multimodal dataset with segment-level annotations for fine-grained hateful video detection. By providing predefined, semantically coherent segment boundaries and specific offensive categories, \textsf{HateClipSeg} addresses critical limitations of prior datasets that relied on coarse video-level labels. Our three-step annotation protocol significantly improves inter-annotator agreement, yielding high-quality labels essential for training robust models. We benchmarked \textsf{HateClipSeg} across three challenging tasks: trimmed video classification, temporal localization, and online classification. Results reveal that while current models can moderately classify trimmed segments, their performance in temporal localization and real-time detection remains limited, underscoring the complexity of segment-level detection in multimodal streams. These findings emphasize the need for new architectures that can better capture multimodal and temporal dependencies in complex, real-world scenarios. 

\textsf{HateClipSeg} offers a valuable resource for advancing research in multimodal hate speech detection. Future work could explore integrating context-aware models, expanding annotations, and addressing emerging challenges such as live content moderation. We will release the dataset and accompanying benchmarks to support further development in this critical area.

\section{Ethical Considerations and Privacy}
This dataset was developed to support research on the detection and understanding of hate speech in online videos, with the aim of mitigating harmful content. Videos were sourced from publicly accessible platforms (YouTube and BitChute) and selected based on their relevance to hate speech targeting protected characteristics such as race, religion, gender, and sexuality. To respect privacy and platform terms, only video IDs are shared, no video content is distributed. Annotators were explicitly warned about the potentially offensive nature of the material before participation and were provided with access to university-supported psychological resources. To protect annotators’ well-being, they were encouraged to take breaks and could skip any video without penalty. Annotations were conducted carefully, with any personally identifiable information (PII) excluded or anonymized. Access to the dataset is restricted to researchers with a legitimate academic or societal interest and is governed by ethical use agreements. This work adheres to institutional ethical guidelines and aims to balance research utility with respect for individuals represented in the data.

\section*{Acknowledgement}
This research / project is supported by A*STAR under its Online Trust and Safety Research Programme (Award Grant No. S24T2TS007), and Ministry of Education, Singapore, under its Academic Research Fund (AcRF) Tier 2. Any opinions, findings and conclusions or recommendations expressed in this material are those of the author(s) and do not reflect the views of the A*STAR and Ministry of Education, Singapore.



\bibliographystyle{ACM-Reference-Format}
\balance
\bibliography{ref}

\end{document}